%% file: main.tex
\documentclass{INTERSPEECH2023}
\usepackage{floatrow}
\usepackage{amsmath,graphicx}
\usepackage[table,xcdraw]{xcolor}
\usepackage{url}
\usepackage{amssymb}
\usepackage{multicol}
\usepackage{multirow}
\usepackage{threeparttable,booktabs}
\usepackage{booktabs}
\usepackage{cite}
\usepackage{capt-of}
\usepackage{bbm}

\usepackage{caption}
\usepackage{subcaption}

\interspeechcameraready 

\title{SparseVSR: Lightweight and Noise Robust Visual Speech Recognition}
\name{Adriana Fernandez-Lopez$^1$, Honglie Chen$^1$, Pingchuan Ma$^{1,2}$, Alexandros Haliassos$^{2}$, Stavros Petridis$^{1,2}$, Maja Pantic$^{1,2}$}
\address{
  $^1$Meta AI, UK\\
  $^2$Imperial College London, UK}
\email{\{afernandezlopez, hongliechen\}@meta.com, \{pingchuan.ma16, alexandros.haliassos14, stavros.petridis04, m.pantic\}@imperial.ac.uk}

\begin{document}

\maketitle
 
\begin{abstract}
Recent advances in deep neural networks have achieved unprecedented success in visual speech recognition. However, there remains substantial disparity between current methods and their deployment in resource-constrained devices. 
In this work, we explore different magnitude-based pruning techniques to generate a lightweight model that achieves higher performance than its dense model equivalent, especially under the presence of visual noise. Our sparse models achieve state-of-the-art results at 10\% sparsity on the LRS3 dataset and outperform the dense equivalent up to 70\% sparsity. We evaluate our 50\% sparse model on 7 different visual noise types and achieve an overall absolute improvement of more than 2\% WER compared to the dense equivalent. Our results confirm that sparse networks are more resistant to noise than dense networks.
\end{abstract}

\input{sections/intro}
\input{sections/method}

\input{sections/experiments}

\input{sections/results}
\input{sections/conclusions}

\section{Acknowledgements}
    Only non-Meta authors conducted any of the dataset pre-processing (no dataset pre-processing took place on Meta’s servers or facilities).

\bibliographystyle{IEEEtran}
\bibliography{mybib}

\end{document}

%% file: sections/intro.tex
\section{Introduction}
\label{sec:intro}

Visual Speech Recognition (VSR) systems decode speech exclusively from video signals based on lip movements. Recently, they shifted from constrained recognition tasks to handle more realistic scenarios, including continuous speech recognition \cite{ma2022visual, shi2022learning, haliassos2022jointly, afouras2018deep, fernandez2018survey}. 
These advances are due to i) the emergence of deep neural networks (DNN) and ii) the availability of large datasets \cite{son2017lip, afouras2018deep}. 
On one hand, it has been shown that deeper models tackle more complex recognition tasks at the cost of heavy networks that are over-parameterised and require a high computational cost and storage resources. On the other hand, they become data-hungry, consisting of very deep architectures with tens or even hundreds of millions of parameters \cite{ma2022visual, ma2021end, ma2023auto, afouras2018deep, ren2021learning, serdyuk2021audio, afouras2020asr}. 
This fact magnifies the gap between research and practice, since many real-life applications demand compact and efficient networks that fulfil different devices. 
In this paper, rather than focusing on investigating larger architectures or datasets, we explore lightweight models while preserving the model's performance.

There has been limited studies on lightweight models in the VSR domain. MobileNet \cite{koumparoulis2019mobilipnet} and EfficientNet \cite{koumparoulis2022accurate} aim to accelerate the inference speed by designing resource-efficient architectures. Ma et al. \cite{ma2021towards} explores distilling the knowledge from a large teacher model to small student models for isolated word recognition.
Our work is more broadly related to model compression, including network pruning \cite{frankle2019lottery, chen2020lottery, ding2021audio, renda2020comparing, frankle2021pruning}, quantization \cite{gui2019model, sainath2020streaming, nguyen2020quantization} and knowledge distillation \cite{hinton2015distilling, li2018compression}.
In particular, network pruning aims to surpass/preserve the dense model performance while pruning out some network weights, leading to a fast and lightweight model. Pruning methods have principally taken two approaches.
\textit{Regularization and gradual pruning} attempt to achieve sparse networks by pruning the model prior to the training~\cite{liu2022unreasonable}. For example, Liu et al. \cite{liu2022unreasonable} recently presented a random pruning method at initialization that achieves similar performance as dense CNN architectures. Another family of methods, namely \textit{retraining}, aims to tackle this problem based on a dense network that has been trained on the target domain. Follow-up pruning and fine-tuning are done to achieve the target sparsity.
Early works studied the \textit{retraining} technique by fine-tuning the pruned model for several epochs at a fixed learning rate to recover accuracy \cite{han2015learning}.
Recently, we find the work on the Lottery Ticket Hypothesis (LTH) \cite{frankle2019lottery} that proposes \textit{weight rewinding}, which sets the unpruned weights back to their initial values and retrains using the original training scheduler. 
Concretely, LTH demonstrated that small-scale networks for computer vision contain sparse matching subnetworks capable of training from initialization to match or even surpass the performance of their dense counterparts.
Following work from Ding et al. \cite{ding2021audio} investigate the possibility of using LTH to discover ASR models that are robust to acoustic noise, transferable, and compatible with structured sparsity. 
Alternatively, Renda et al. \cite{renda2020comparing} proposed \textit{learning rate rewinding} (LRR), which sets the unpruned weights to their final values and retrains from there using the original learning rate scheduler. In another direction, Lai et al. \cite{lai2021parp} present Prune-Adjust-Re-Prune (PARP) for learning a self-supervised speech representation, where promising results are shown at a single fine-tuning run by allowing regrowth of their weights after pruning.

In this work, we consider the problem of pruning VSR networks. We explicitly look for a lightweight model computationally and memory efficient, transferable, and noise-robust. To this end, we conduct an analysis of various network pruning methods and set the first sparse model benchmark in VSR. 
Concretely, our contributions are four-fold: (i), we perform an ablation study to identify the best rewinding option for VSR models. Since several studies suggest that for large-scale settings \cite{frankle2020linear, chen2020lottery} ($>$25 M parameters) it is difficult to recover from initial weights and there is a need to initialize early in training, we extensively analyse the capacity of a sparse model when rewinded at different epochs. Similarly to Renda et al. \cite{renda2020comparing}, we observe that for VSR models the better we chose the rewinding weights, the luckiest the model recovers the loss of information from pruning, especially at high levels of sparsity;
(ii), we compare the behaviour of different magnitude-based pruning techniques recently applied to ASR and NLP \cite{chen2020lottery, lai2021parp, ding2021audio}. Consistently with the ablation study, we observe that iterative pruning with rewinding to the final epoch provides the best results and significantly surpasses the accuracy of the dense BASE-S model by 1\% Word Error Rate (WER) absolute improvement at 50\% sparsity. 
(iii), we compare our sparse models to state-of-the-art VSR models \cite{shi2022learning, ma2023auto}. On one hand, and differently from previous works, we introduce visual noise augmentations during training to push forward the achievable performance of VSR models. This prevents over-fitting, which is especially important for training sparse models that are trained over multiple rounds. Thus, our noise-exposed sparse BASE-L models reach state-of-the-art results on LRS3 with 19.5\% WER at 10\% sparsity and outperform the dense model up to 70\% sparsity; 
(iv), since noise robustness is a crucially demanded technological factor for VSR models to work in real scenarios, we conduct an analysis of 7 different visual noise types and show up more than 2\% WER absolute improvement of 50\% sparse models with respect to the dense model, suggesting that sparse models are more robust to noise than the dense counterpart. 

%% file: sections/method.tex
\section{Approach}
\label{sec:pruning_methods}
\subsection{Dense architecture}
\label{sec:dense_model}
Given a variable length video sequence $X\in\mathbb{R}^{3 \times N \times h \times w}$ with $N$ frames of shape $h \times w \times 3$, we compute its corresponding grapheme transcription $Y\in\mathbb{R}^{1 \times L}$ (with $L$ graphemes) using a function $f(\cdot;\theta_T)$, where $\theta_T\in \mathbb{R}^p$ conform the model parameters, $p$ is the number of parameters of the network, $T$ is the number of training epochs and $\theta_0$ are its initial weights. In this case, we define the dense network in eq. (\ref{eq:dense_model}) as a Sequence-to-Sequence (S2S) + Connectionist Temporal Classification (CTC) model that mainly consists of a CNN frontend, a Conformer encoder and a Transformer decoder~\cite{ma2022visual}.
\begin{align}
    Y = f(X;\theta_T), \quad Y \in \mathbb{R}^{1 \times L }
\label{eq:dense_model}
\end{align}
\subsection{Subnetwork discovery}
\label{sec:sparse_model}
To construct a sparse subnetwork, we follow the standard pruning approach that removes collectively the weights with the lowest magnitude, independently of their location in the dense network, namely, unstructured global magnitude pruning~\cite{han2015learning}. Particularly, once the dense model has been trained for $T$ epochs, the parameters of the dense model $\theta_T$ are pruned using a binary mask $m\in\{0,1\}^p$ to meet the target sparsity, i.e. $\theta'=m\odot\theta_T$,
where $\odot$ is the element-wise product, $p$ is the number of network parameters and $\theta'$ are the model parameters after pruning. 
The goal is to learn $(m, \theta')$ such that the sparse subnetwork $f(\cdot; \theta')$ achieves minimal fine-tuned loss $\mathcal{L}$ on the training data. 



\subsubsection{Na\"ive pruning} 
The most straightforward approach is to mask out the pretrained dense model to the target sparsity and keep the non-zero parameters unchanged, as shown in eq. (\ref{eq:naive}). Where $\theta_T$ are the parameters of the pretrained dense model. This model allows no further fine-tuning and is agnostic to downstream tasks. 
\begin{align}
   f(\cdot;\theta') = f(\cdot ; m\odot\theta_T)
\label{eq:naive}
\end{align}

\subsubsection{One-shot pruning} 
\label{sec:one_shot}

Na\"ively pruning the dense model is computationally efficient, however, it also prevents the model learning from additional signals.
In \textit{one-shot pruning}, 
we take additional steps by fine-tuning or rewinding on downstream datasets.
Specifically, we generate the mask $m$ by pruning the parameters of the pretrained dense model $\theta_T$. To create an initial subnetwork, the model state $\theta_t$ with $t \in [0,T]$ is sparsified using the mask $m$.
Finally, the non-zero weights are trained on the downstream datasets to reach the dense performance.
Formally, given a training dataset $\mathcal{D} = \{d_{1}, d_{2},..., d_{k}\}$, the algorithm is formulated as eq. (\ref{eq:LTH}). Where $\mathcal{A}$ denotes an optimization algorithm (e.g. Adam) and $i$ refers to the number of training iterations.
Note, the $\textsc{Frozen}$ operation only freezes the pruned weights and keeps other weights adjustable. 
\begin{align}
   f(\cdot;\theta') = \mathcal{A}_{i}^{D}(f(\cdot;\textsc{Frozen}(m\odot\theta_t
   )))
   \label{eq:LTH}
\end{align}  
We follow the previous works and compare $4$ variants of non-zero weights recovery methods. 
First, models are fine-tuned using a constant learning rate \cite{han2015learning}; 
Second, following PARP \cite{lai2021parp}, weights are allowed to regrow during fine-tuning and the target sparsity is achieved with an additional na\"ively pruning;
Third, following LTH \cite{frankle2019lottery}, we rewind the weights and learning rate to their initial values ($t=0$);
Fourth, following LRR \cite{renda2020comparing}, we only rewind the learning rate to their initial values and initialise the weights using the dense model ($t=T$).
We explicitly analyse the choice of $t$ and identify the best pruning method for VSR.

\subsubsection{Iterative pruning} 
\textit{Iterative pruning} was proposed \cite{han2015learning, frankle2019lottery, lai2021parp}, where the subnetwork discovery phase is split into multiple rounds. Thus, \textit{iterative pruning} is able to extract better subnetworks than one-shot, especially at high sparsity levels, but requiring a huge computational cost due to their iterative training procedure. Concretely, the model is iteratively pruned and retrained for further $i$ iterations during $r$ rounds to reach the target sparsity. Following the procedure described in \cite{chen2020lottery}, each round prunes $1/r$ of the non-zero weights. After pruning each round, the model parameters can be rewinded to their initial weights $\theta_0$, their intermediate weights $\theta_{t}$ or their final weights $\theta_T$. In this case, 
we perform $r=10$ rounds with a weight drop of $10\%$ every time. 




%% file: sections/experiments.tex
\section{Experiments}
\label{sec:settings}

\subsection{Dataset}
We conduct experiments on the LRS3 dataset \cite{afouras2018lrs3}, which is the largest publicly available audiovisual dataset in English collected from TED and TEDx talks. It contains 150,498 utterances for training (438 hours) and 1,321 utterances for testing (0.9 hours). 
Furthermore, we use additional audiovisual data such as VoxCeleb2 \cite{chung2018voxceleb2} and AVSpeech \cite{ephrat2018looking} as the training data. In particular, we use VoxLingua107 language classifier to filter English-speaking videos and then take advantage of the Wav2Vec2.0 \cite{baevski2020wav2vec} to automatically annotate the videos, resulting in a total of 1,307 and 1,323 hours, respectively.

\subsection{Data augmentation}
\label{sec:DA}
We apply horizontal flipping, random cropping, and adaptive time masking \cite{ma2022visual} to all our models. We choose masks that are proportional to the utterance length and a maximum masking length of up to 0.4 seconds.
Furthermore, we also provide the possibility of visual noise augmentations. We follow the corruption pipeline shown in \cite{haliassos2021lips} and generate 10 levels of increasing corruption for the noise types: block-wise distortions (BW), Gaussian blur (GB), motion blur (MB) and pixelation (P). 
For each training step, a noise type as well as the clean data were selected using a uniform distribution. Following the selection of the noise type, the noise level was randomly determined.

\subsection{Implementation details}
\label{sec:experimental_setup}
\par{\noindent \bf Pre-processing.} We follow the pre-processing steps in prior works \cite{ma2022visual, shi2022learning} to crop a 96$\times$96 region centred around the mouth, afterwards each frame is transformed into a gray level.

\par{\noindent \bf Architecture details.} The proposed VSR baseline follows the architecture in \cite{ma2022visual}. We set up a BASE-S model to perform ablation studies and a BASE-L model to provide competitive results on VSR. Both models consist of a 3D convolutional layer with a receptive field of 5 frames, followed by a 2D ResNet-18 \cite{petridis2018end}, a Conformer encoder, a projection CTC layer and a Transformer decoder. For BASE-S (BASE-L), we use a 12-layer Conformer encoder with 256 (768) input dimensions, 2,048 (3,072) feed-forward dimensions, and 4 (12) attention heads. Each model decoder is a 6-layer Transformer with the same dimensions and number of heads as the corresponding encoder, resulting in 56.4 M (250.4 M) parameters for BASE-S (BASE-L) models. 

\par{\noindent \bf Training strategy.} The ResNet-18 is initialized with the available model from \cite{ma2022visual}. Finally, we train using a combination of CTC loss and Cross-Entropy loss. The network output is decoded using 5,000 unigram subwords \cite{ma2023auto}. 
Our models use AdamW \cite{loshchilov2019decoupled} optimizer, with $\beta_1=0.9$, $\beta_2=0.98$, a cosine learning rate scheduler, and a peak learning rate of 6e-4 (1e-3) for BASE-S (BASE-L). 
We warm up the first $5$ epochs and train for $75$ epochs. Each round of iterative pruning of sparse models takes $50$ epochs. The maximum number of frames in each batch is 2,400 (1,800) for BASE-S (BASE-L) models. 
On $16$ ($64$) A$100$ GPUs, the training takes around $10$ ($60$) hours for BASE-S (BASE-L) models. For comparison purposes, no language model (LM) is incorporated in this work.
\par{\noindent \bf Pruning details.} We follow previous work \cite{frankle2019lottery} and prune weights from linear, 1D-, 2D- and 3D-convolutional layers.

%% file: sections/results.tex
\section{Results}
\label{sec:results}


\subsection{Ablation study on rewinding options}

\input{tables/Fig-ablation_study}

\input{tables/Tab-pruning_methods}

To identify the optimal rewinding weights for generating a sparse VSR model, we perform an ablation study. We follow one-shot pruning, which is computationally more efficient than iterative pruning for this study. Initially, we train a dense BASE-S model on the LRS3 dataset. Afterward, we look for three subnetworks at sparsity levels 20\%, 50\% and 80\% that report the model's performance. We analyse the recovery capacity of each network when rewinded at different epochs $t$, where $t=$\{0, 5, 10, 20, 40, 60, 75\}. To this end, we prune the dense model to target sparsity, set up the unpruned weights to previous values $\theta_t$ and reset the learning scheduler for retraining.  

Figure~\ref{fig:ablation_study} shows the results. We observe that for 20\% and 50\% sparsity levels, the model remains stable for any initialization weights, i.e. the model is able to recover dense performance with very small degradation. However, the performance of the 20\% sparse model outperforms the dense model by 2\% WER absolute difference only when rewinded from the last training weights. Furthermore, and consistent with previous observations \cite{frankle2020linear, renda2020comparing}, for a large setup (more than 25M parameters) and a large sparsity level, the proper choice of the rewinding weights have an impact on the recovery loss. For example, the 80\% sparse model reports more than 5\% WER absolute improvement once the rewinding weights correspond to the last training epoch $t=T$ compared to the initial weights $t=0$. 
Therefore, we conduct the rest of the analysis by rewinding to the last training values and resetting the learning rate scheduler \cite{renda2020comparing}.

\subsection{Comparison of network pruning techniques}
\label{sec:results_pruning}

In this section, we investigate the impact of several pruning techniques on the performance on the LRS3 test set. In particular, we compress our dense BASE-S model by using na\"ive pruning, 4 one-shot pruning, and 2 iterative pruning algorithms at different sparsity levels from 0\% to 90\%, with an interval step of 10\%. Results are shown in Table \ref{tab:comparison}. We observe that as long as the sparsity level remains below 20\% our sparse models outperform the dense model in almost all cases, which indicates that VSR models are over-parameterized. However, as the sparsity value increases, the performance after vanilla pruning gets compromised. For example, when 50\% of the parameters are removed, there is an increase of 22.8\% in WER. This is most likely due to the vast amount of parameters lost in the model, which makes the recovery procedure especially challenging. 
On the other hand, by applying iterative pruning \textit{LRR} \cite{renda2020comparing}, we observe an absolute improvement of 1-2\% WER compared to the dense model from 10\% to 50\% sparsity. Similarly to findings in \cite{renda2020comparing}, this indicates that such pruning methodology seems like a silver bullet for reducing the issue of overfitting. 

\input{tables/Tab-SOTA_and_other_compression}
\subsection{Lightweight competitive VSR models}
\label{sec:large_model}

In Table \ref{tab:SOTA}, we show the performance of dense and sparse models on the LRS3 test set. We observe that our dense models are comparable to similar work on VSR, using less training data. We highlight the benefits of adding noise during training, which provides data augmentation that allows the model to generalize better, which is especially important for training sparse models over multiple rounds. Next, based on our dense BASE-L model exposed to noise, we iteratively generate the sparse models. We observe that our sparse models reach state-of-the-art results with 19.5\% WER at 10\% sparsity. Additionally, they outperform the dense model up to 70\% sparsity and remain without significant degradation until 80\% sparsity (where the dense BASE-S model with a similar amount of parameters performs worse on the same setup, 21.8\% WER vs 26.6\% WER). 
Furthermore, we also highlight the benefits of sparse networks against other compression techniques. Table \ref{tab:SOTA} shows the performance of the dense BASE-L model exposed to noise when trained for additional $75$ epochs on non-uniform 8-bit quantization-aware-training, following a well-established method from \cite{gui2019model}. Concretely, we observe that the quantized model does not suffer degradation, but does not provide any benefits in terms of performance. On the other hand, we follow Ma et al. \cite{ma2021towards} to distill BASE-S student networks that share a similar amount of parameters as our sparser network without significant accuracy loss (i.e. 80\% sparsity) using our dense BASE-L model as a teacher. We observe that the 80\% sparse model outperforms the best distilled network by 5\% WER absolute difference. 

From these results, we claim that our sparse networks are able to highly compress dense networks up to 3-4 times without accuracy loss, which pushes forward the state-of-the-art results while at the same time reducing model size.

\subsection{Noise experiments}
\label{sec:noise_robustness}

\input{tables/Tab-Fig-50_noises}

In order to investigate the robustness of our VSR models against visual noise, we run our 50\% sparse and dense BASE-L models at different levels of noise. Note that all models are augmented with 4 visual noises (BW, GB, MB and P) and 10 levels of corruption as well as clean image sequences. Results are shown in Figures \ref{fig:seen_noises} and \ref{fig:unseen_noises}. Overall, we show that the gap between the sparse model and the dense model becomes increasingly larger as the level of noise increases, which indicates that the sparse model is more beneficial when image sequences are heavily corrupted. The same conclusions can be drawn when unseen noise such as JPEG video compression (VC), Gaussian noise (GN) and contrast (C) is added to the test set. 
In particular, when GN is added to the test set, our sparse model outperforms the dense model by 14.7\% WER absolute improvement at level 5. 
This is probably because the sparse model filters deeper the image and only retains the relevant details, while the dense model may consider the noise. On the other hand, we show a 1.4\% improvement on VC at level 1, which reduces as the level of corruption increases, indicating that our VSR models are less sensitive to video compression noise. This is likely due to the loss of important details on the mouth and lip appearance at high compression ratios of that noise type.

%% file: tables/Fig-ablation_study.tex

\begin{figure}[!tb]
  \centering
  \vspace{-0.2cm}
  \includegraphics[width=0.9\columnwidth]{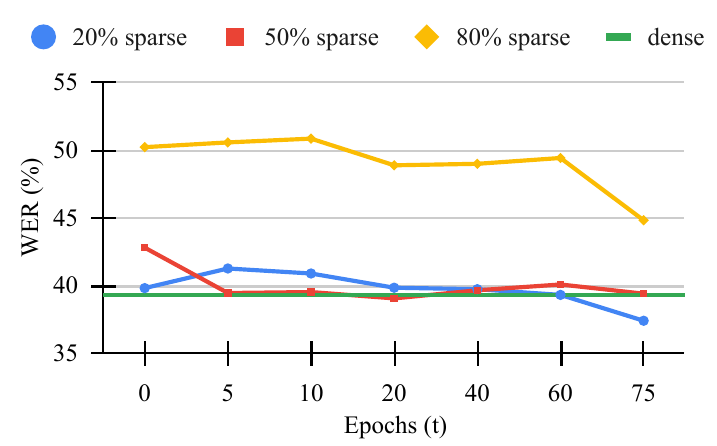}
  \vspace{-0.2cm}
  \caption{WER of 20\%, 50\% and 80\% sparse BASE-S models evaluated on LRS3 when rewinded at different epochs $t$.} 
  \label{fig:ablation_study}
\end{figure}

%% file: tables/Tab-pruning_methods.tex
\begin{table}[!htb]
\captionof{table}{WER (\%) of different pruning techniques evaluated on  LRS3. $Sp$ means the sparsity level. Models do not use LM.}
\vspace{-.2cm}
\footnotesize
\setlength{\tabcolsep}{2.5pt}
\centering
\begin{threeparttable}
\begin{tabular}{l c c c c c c c}
\toprule
 &       Na\"ive                   & \multicolumn{4}{c}{One-shot pruning}                                & \multicolumn{2}{c}{Iterative pruning}   \\
\cmidrule(lr){3-6} 
\cmidrule(lr){7-8}
$Sp$             &  Pruning          & Finetune$^*$ & PARP$^*$           & LTH            & LRR           & LTH            & LRR           \\ \midrule
0                    & 39.30          & 39.30    & 39.30          & 39.30          & 39.30          & 39.30          & 39.30          \\
10\%                 & 39.20         & 40.15    & 39.36          & 39.11           & 37.40          & 40.78          & 37.88          \\
20\%                 & 39.80          & 40.23    & 39.25         & 39.80          & 37.39           & 40.78          & 37.30          \\
30\%                 & 40.80          & 40.65    & 39.63          & 40.00          & 38.38          & 40.78          & 38.38          \\
40\%                 & 44.50          & 41.11    & 40.30          & 40.91          & 38.55          & 40.78          & 37.80          \\
50\%                 & 62.10          & 43.04    & 45.30          & 42.79          & 39.39          & 40.78           & 38.26         \\
60\%                 & 122.3          & 47.67    & 61.56          & 43.89          & 40.23          & 42.34          & 40.13          \\
70\%                 & 106.1          & 59.89    & 92.05          & 45.50          & 42.26          & 42.34          & 41.29          \\
80\%                 & 110.8          & 95.99    & 376.00          & 50.21          & 44.81          & 48.40          & 43.53          \\
90\%                 & 100.0          & 100.0    & 141.00          & 68.45          & 64.40          & 57.69          & 53.12         \\
\bottomrule 
\end{tabular}
\begin{tablenotes}
\item[*]Finetune for 10 epochs with a fixed learning rate of 1e-5, similar to the last value in training.
\end{tablenotes}
\end{threeparttable}
\label{tab:comparison}
\vspace{-0.2cm}
\end{table}

%% file: tables/Tab-SOTA_and_other_compression.tex
\begin{table}[!htb]
\caption{Results of VSR networks on LRS3. Sparse models are generated by iterative pruning LRR. Our models do not use LM.}
\vspace{-.2cm}
\label{tab:SOTA}
\setlength{\tabcolsep}{1.8pt}
\centering
\begin{threeparttable}
\begin{tabular}{lcccr}
\toprule
\multirow{2}{*}{\textbf{Method}} & \multirow{2}{*}{\begin{tabular}[c]{@{}c@{}}\textbf{Training} \\ \textbf{Data (hours)}\end{tabular}}          & \multirow{2}{*}{\textbf{Noise}} & \multirow{2}{*}{\begin{tabular}[c]{@{}c@{}}\textbf{WER} \\ \textbf{(\%)}\end{tabular}}  & \multirow{2}{*}{\textbf{\# Params.}} \\ 
 & & & & \\ \midrule
CM-seq2seq \cite{ma2021end} &  438 & \text{\sffamily X} & 46.9 & 56.4 M \\
KD+CTC \cite{afouras2020asr} & 772 & \text{\sffamily X}  & 59.8 & 212 M \\
AVHuBERT \cite{shi2022learning} &            1,759 &  \text{\sffamily X}  & 26.9 & 325 M \\
RAVEn \cite{haliassos2022jointly} &            1,759 &  \text{\sffamily X}  & 24.4 & 493 M \\
TM-seq2seq \cite{afouras2018deep} & 1,362 & \text{\sffamily X}  & 58.9 & 54.2 M \\
Auto-AVSR \cite{ma2023auto}      & 3,448                 & \text{\sffamily X}                       & 20.5              & 250.4 M             \\ 
RNN-T \cite{makino2019recurrent} & 31,000 & \text{\sffamily X}  & 33.6 & 62.9 M \\ \midrule
dense BASE-S    & \multirow{3}{*}{438} & \text{\sffamily X}                                & 39.3              & 56.4 M              \\
20\% sparse BASE-S  &                         & \text{\sffamily X}                             & 37.3     & 45.4 M             \\ 
50\% sparse BASE-S  &                         & \text{\sffamily X}                             & 38.2     & 28.9 M             \\ \midrule
dense BASE-S    & \multirow{12}{*}{3,068\tnote{$*$}}                          & \checkmark                               & 26.6              & 56.4 M              \\
dense BASE-L    &                         & \text{\sffamily X}                               & 21.1              & 250.4 M             \\
dense BASE-L    &                         & \checkmark                               & 20.3     & 250.4 M             \\
10\% sparse BASE-L  &                         & \checkmark                               & 19.5     & 225.7 M             \\
20\% sparse BASE-L  &                         & \checkmark                               & 20.0     & 201.1 M             \\
30\% sparse BASE-L  &                         & \checkmark                               & 19.8    & 176.5 M             \\
40\% sparse BASE-L  &                         & \checkmark                               & 20.2     & 151.9 M             \\
50\% sparse BASE-L  &                         & \checkmark                               & 19.6     & 127.3 M             \\
60\% sparse BASE-L  &                         & \checkmark                               & 19.9    & 102.7 M             \\
70\% sparse BASE-L  &                         & \checkmark                               & 20.1     & 78.1 M              \\
80\% sparse BASE-L  &                         & \checkmark                               & 21.8              & 53.5 M              \\
90\% sparse BASE-L  &                         & \checkmark                               & 26.3             & 28.8 M              \\ \midrule
8-bit dense BASE-L   &    \multirow{3}{*}{3,068\tnote{$*$}}                    & \checkmark & 20.3                  & 250.4 M                                \\
distilled BASE-S        &                         & \text{\sffamily X} &  26.8                 & 56.4 M                                \\
distilled BASE-S        &                         & \checkmark &  32.1                 & 56.4 M                                \\
\bottomrule
\end{tabular}
\begin{tablenotes}
\item[$*$] LRS3, VoxCeleb2 and AVSpeech
\vspace{-.3cm}
\end{tablenotes}
\end{threeparttable}
\end{table}

%% file: tables/Tab-Fig-50_noises.tex



\begin{figure}[!tb]
\centering
\begin{subfigure}[b]{0.85\textwidth}
   \includegraphics[width=\columnwidth, trim=0.2cm 0.1cm 1cm 0cm, clip]{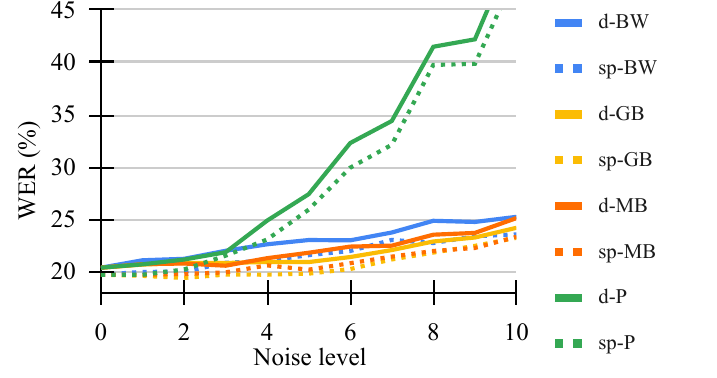}
   \caption{Seen noise types.}
   \label{fig:seen_noises}
\end{subfigure}
\vspace{-0.1cm}
\begin{subfigure}[b]{0.85\textwidth}
   \includegraphics[width=\columnwidth, trim=0.2cm 0cm 1cm 0.2cm, clip]{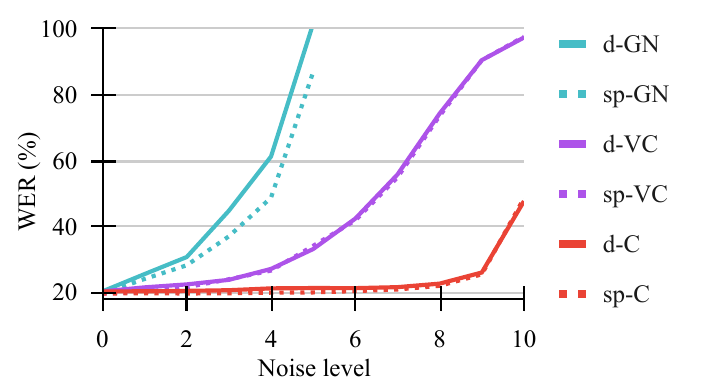}
   \caption{Unseen noise types.}
   \label{fig:unseen_noises}
\end{subfigure}
\vspace{-0.1cm}
\caption[]{WER of noise-exposed dense "d" and 50\% sparse "sp" BASE-L models on the LRS3 test set corrupted by visual noises at different levels (level 0 means clean data).}
\end{figure}

%% file: sections/conclusions.tex
\section{Conclusions}
\label{sec:conclusion}

In this paper, we revisit network pruning techniques and set the first benchmark in a completely new domain, VSR. 
For this task, we conduct thorough ablation studies to validate the best rewinding option and experiment with different pruning methods.
We present state-of-the-art results on the LRS3 dataset using VSR models at different sparsity levels. We also show no degradation up to 70\% sparsity and a model parameter reduction rate up to 3.2.
Finally, we reinforce the fact that under similar conditions, sparse networks are more robust against noise than their dense counterpart. Future work may investigate the combination of multiple compression methods.